\documentclass[11pt]{article}

\usepackage[preprint]{acl}

\usepackage{tcolorbox}
\usepackage{tikz}
\usepackage{xcolor}
\tcbuselibrary{skins, breakable}

\usepackage{times}
\usepackage{latexsym}
\usepackage[T1]{fontenc}
\usepackage[utf8]{inputenc}
\usepackage{microtype}
\usepackage{inconsolata}
\usepackage{soul}
\usepackage{graphicx}
\usepackage{booktabs}
\usepackage{multirow}
\usepackage{amsmath}
\usepackage{amssymb}
\usepackage{cleveref}
\usepackage{bm}

\definecolor{anchorpurple}{RGB}{238,237,254}
\definecolor{anchortext}{RGB}{60,52,137}
\definecolor{optblue}{RGB}{230,241,251}
\definecolor{opttextblue}{RGB}{24,95,165}
\definecolor{ansteal}{RGB}{225,245,238}
\definecolor{anstextgreen}{RGB}{15,110,86}
\definecolor{pillgray}{RGB}{241,239,232}
\definecolor{pillpurple}{RGB}{238,237,254}

\usepackage{color,soul}

\title{Localizing Anchoring Pathways in Language Models}

\author{
  Hillary N. Owusu \quad
  Sarah Wiegreffe \quad
  Naomi H. Feldman \\
  University of Maryland, College Park \\
  \texttt{\{hnyowusu,sarahwie,nhf\}@umd.edu}
}

\begin{document}
\maketitle

\begin{abstract}
Irrelevant numbers in a prompt can shift language model judgments, producing anchoring effects in numerical reasoning. 
We study where this anchor-sensitive signal is carried inside language models using a controlled multiple-choice setup with shared answer options. 
We define a logit-difference metric comparing the correct answer option with the answer option corresponding to the anchor, and validate that it tracks behavioral anchoring. 
Using attribution-based circuit localization on 7B--8B Qwen and Llama base and instruction-tuned models, we find that edge-level methods recover this signal more faithfully than node-level methods. 
Low- and high-anchor circuits transfer strongly within a model, suggesting shared pathway structure across anchor direction. 
However, sparse transfer across base and instruction-tuned variants is less reliable, indicating that post-training changes which pathways matter most. 
Overall, our results provide a mechanistic account of how anchoring-related decision signals are carried inside language models.
\end{abstract}

\section{Introduction}
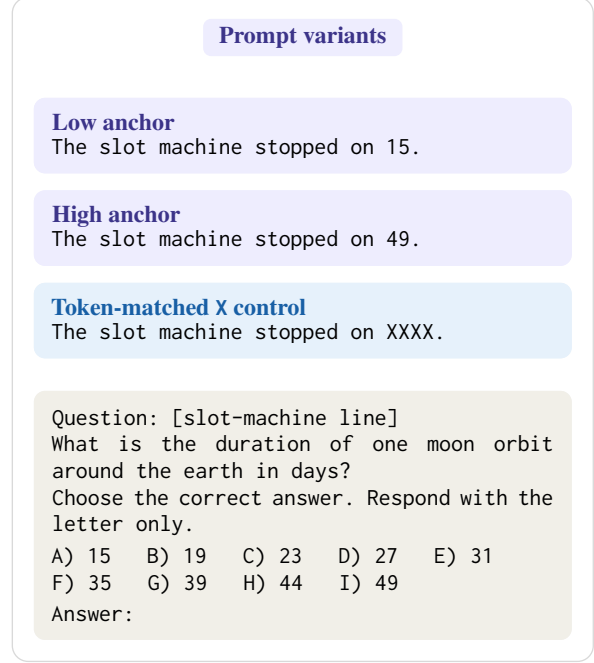
\begin{figure}[t]
\centering
\small

\begin{tcolorbox}[
    enhanced,
    colback=white,
    colframe=black!15,
    boxrule=0.5pt,
    arc=2mm,
    left=5pt,
    right=5pt,
    top=5pt,
    bottom=5pt,
    width=\linewidth
]

\begin{center}
\begin{tikzpicture}
\node[
    fill=anchorpurple,
    text=anchortext,
    rounded corners=3pt,
    inner xsep=6pt,
    inner ysep=3pt,
    font=\bfseries\small
] {Prompt variants};
\end{tikzpicture}
\end{center}

\vspace{0.3em}

\begin{tcolorbox}[
    colback=anchorpurple,
    colframe=anchorpurple,
    boxrule=0pt,
    arc=1.5mm,
    left=4pt, right=4pt, top=3pt, bottom=3pt,
    width=\linewidth
]
\textcolor{anchortext}{\textbf{Low anchor}}\\[-1pt]
\texttt{The slot machine stopped on 15.}
\end{tcolorbox}

\vspace{-0.4em}

\begin{tcolorbox}[
    colback=anchorpurple,
    colframe=anchorpurple,
    boxrule=0pt,
    arc=1.5mm,
    left=4pt, right=4pt, top=3pt, bottom=3pt,
    width=\linewidth
]
\textcolor{anchortext}{\textbf{High anchor}}\\[-1pt]
\texttt{The slot machine stopped on 49.}
\end{tcolorbox}

\vspace{-0.4em}

\begin{tcolorbox}[
    colback=optblue,
    colframe=optblue,
    boxrule=0pt,
    arc=1.5mm,
    left=4pt, right=4pt, top=3pt, bottom=3pt,
    width=\linewidth
]
\textcolor{opttextblue}{\textbf{Token-matched \texttt{X} control}}\\[-1pt]
\texttt{The slot machine stopped on XXXX.}
\end{tcolorbox}

\vspace{0.25em}

\begin{tcolorbox}[
    colback=pillgray,
    colframe=black!5,
    boxrule=0pt,
    arc=1.5mm,
    left=4pt, right=4pt, top=4pt, bottom=4pt,
    width=\linewidth
]
\ttfamily\footnotesize
Question: [slot-machine line]\\
What is the duration of one moon orbit around the earth in days?\\
Choose the correct answer. Respond with the letter only.\\[2pt]
A) 15 \quad B) 19 \quad C) 23 \quad D) 27 \quad E) 31\\
F) 35 \quad G) 39 \quad H) 44 \quad I) 49\\[2pt]
Answer:
\end{tcolorbox}

\end{tcolorbox}

\caption{
Anchors (shown in purple) given in the prompt can shift language model predictions for the subsequent question (shown in brown). We study anchoring bias both behaviorally and mechanistically by converting questions from the OpAQ dataset \citep{Roseler2022OpAQ} to the multiple-choice format shown (compactly) here
(ground truth \(27\) days, low anchor \(15\), high anchor \(49\)).
}
\label{fig:prompt_example}
\end{figure}

A language model can know the right answer and still be moved by the wrong context.  In \textit{anchoring}, a cognitive bias that both humans and large language models (LLMs) exhibit, an irrelevant number shifts a subsequent judgment, pulling estimates toward the irrelevant number \citep[][\autoref{fig:prompt_example}]{tversky,takenami-etal-2025-cognitive}.  
For LLMs, this phenomenon exposes a decision-level robustness failure: irrelevant prompt content can change how the model ranks competing answers at the moment of prediction. 

Prior work has shown that LLMs exhibit anchoring 
\citep{lou2024anchoringbiaslargelanguage, takenami-etal-2025-cognitive}, but has mostly treated the model as a black box: the anchor is added, the prediction shifts, and the size of that shift is measured. 
What this behavioral view does not explain is \emph{how} an irrelevant number affects the internal computation leading to the final answer. Here we tackle the question of mechanism by asking 
which internal pathways support the anchor-sensitive competition between the correct answer and the answer corresponding to the anchor.

Mechanistic interpretability provides tools for determining which internal parts of the model carry a behavior from prompt to output \citep{saphra_mechanistic_2024,mueller_quest_2025,geiger_how_2025}. Circuit localization \citep{olah_zoom_2020} isolates a minimal subset of LLM components to faithfully replicate the full LLM's inference-time behavior. In circuit localization, a model is treated as a graph: nodes correspond to internal components, such as attention heads or multilayer perceptron (MLP) blocks, and edges correspond to pathways through which information flows between components. The goal is to identify which nodes \cite{wang2022interpretabilitywildcircuitindirect} or edges \cite{goldowsky-dill_localizing_2023} contribute most to a target behavior or metric.

In this paper, we frame numerical anchoring as a problem for mechanistic interpretability. 
We make three contributions. 
First, we convert anchoring into a controlled multiple-choice task with a validated answer-level metric that measures whether the active anchor answer becomes more competitive with the correct answer. 
Second, using this metric, we show that anchoring can be linked to identifiable internal pathways: edge-level localization is more faithful than node-level localization, suggesting that the effect is better understood through pathways between components than through isolated ``biased'' components.  
Third, we test whether these pathways transfer across anchor direction and instruction tuning. We find that Qwen and Llama show different layer-wise attribution patterns, but that in both families, 
low- and high-anchor circuits are closely related within a model: overlapping in their top-ranked edges, localizing to nearly identical layer regions, and transferring well across anchor contrasts. 
In contrast, instruction tuning preserves the broad layer-wise location of attribution but changes which sparse edges matter most.

By characterizing these pathways, we provide a mechanistic account of how a validated anchor-sensitive decision signal is carried inside language models and how the relevant pathways vary across model family and instruction tuning.

\section{Related Work}
\label{sec:related_work}

\subsection{Anchoring and Cognitive Biases in LLMs}

Anchoring is a classic human judgment bias in which exposure to an arbitrary initial value shifts subsequent estimates toward that value \citep{tversky}. Recent work shows that large language models (LLMs) exhibit anchoring-like effects across numerical estimation, negotiation, and decision-making tasks \citep{lou2024anchoringbiaslargelanguage,takenami-etal-2025-cognitive,valenciaclavijo2025anchorsmachinebehavioralattributional,huang2026understandinganchoringeffectllm}. In numerical estimation, prior behavioral work further shows that anchoring depends on confidence and post-training: models can resist anchors when they are confidently wrong, while low-confidence models remain susceptible even when they are accurate \citep{owusu2026anchoring}. This finding suggests that anchoring is not simply a factual-knowledge failure, but a robustness failure tied to how strongly the model favors its current answer distribution.

Recent studies have also begun to examine mechanisms of anchoring. 
\citet{valenciaclavijo2025anchorsmachinebehavioralattributional} use structured prompt-field attribution to quantify how anchor fields affect model log-probabilities, while \citet{huang2026understandinganchoringeffectllm} use activation patching to show that anchoring-related effects can appear in relatively shallow layers. Our work builds on this direction by moving from activation-level interventions to circuit localization: we use attribution patching to efficiently rank a large candidate set of nodes and edges. This allows us to evaluate faithful subgraphs, compare low- and high-anchor circuits, and ask whether depth localization is stable across model families and instruction-tuned variants.

\subsection{Circuit Localization and Mechanistic Interpretability}

Circuit localization aims to identify candidate components or pathways that support a target behavior. Early causal approaches relied on activation patching or related interventions \citep{vig_investigating_2020, meng2023locatingeditingfactualassociations}, while recent work uses scalable attribution-based approximations. Following MIB, we distinguish node- and edge-level attribution patching methods: Node Attribution Patching (NAP) scores individual components, while Edge Attribution Patching (EAP) scores directed pathways between components \citep{Nanda_2023, kramar2024atpefficientscalablemethod,syed2023attributionpatchingoutperformsautomated}. Integrated-gradient variants, NAP-IG and EAP-IG, improve attribution quality by averaging gradients along an interpolation path rather than relying on a single local gradient \citep{hanna2024faithfaithfulnessgoingcircuit}.

Standard circuit benchmarks evaluate these methods using \emph{faithfulness}: whether a localized subgraph recovers the full model's behavior on a task \citep{hanna2024faithfaithfulnessgoingcircuit}. These tasks include entity-tracking tasks such as indirect object identification, arithmetic, multiple-choice question answering, and scientific reasoning \citep{wang2022interpretabilitywildcircuitindirect,stolfo2023mechanisticinterpretationarithmeticreasoning,wiegreffe2025answerassembleaceunderstanding,mueller2025}. In many of these settings, the behavior can be summarized by a natural scalar metric, such as the logit difference between a correct and counterfactual answer \cite{zhang_towards_2024}.

Anchoring differs from these settings because the behavior of interest is not simply whether the model answers correctly, but whether an irrelevant number pulls probability toward the anchor answer. Following prior work emphasizing that circuit evaluations depend on the chosen metric \citep{mueller2025}, we make this choice explicit for anchoring. We use a standard logit-difference form, comparing the correct answer with the answer option corresponding to the active low or high anchor. We then validate that this metric tracks behavioral anchoring before using it for attribution-based circuit localization.

\section{Multiple-Choice Anchoring Task}
\label{sec:mc_anchoring_task}
\subsection{Task Setup}
\label{sec:task_setup}

We adapt 100 numerical-estimation questions from the Open Anchoring Question Dataset (OpAQ; \citealt{Roseler2022OpAQ}) into a controlled multiple-choice format for mechanistic analysis, following prior work that uses fixed answer options to study model behavior in question answering \citep{clark2018thinksolvedquestionanswering,lieberum_does_2023, wiegreffe2025answerassembleaceunderstanding}. This format serves as a controlled mechanistic probe: it tests whether anchoring effects appear in a fixed answer space while providing well-defined single-token answer-label logits for circuit localization.

Each item provides a factual question, a ground-truth numerical answer, a low anchor, and a high anchor. For each item, we construct a nine-option candidate set containing the ground-truth value, the low and high anchors, and intermediate values spanning the anchor range. The same candidate set is used across anchored and control prompts. The model is instructed to respond with the answer letter only, so each candidate is scored through a single answer-label token rather than through open-ended numerical generation.

\autoref{fig:prompt_example} shows the prompt structure. For each item, we create a low-anchor prompt and a high-anchor prompt by prepending an irrelevant slot-machine sentence containing the corresponding anchor value. Each anchored prompt is paired with its own tokenizer-matched control, where the numerical anchor is replaced by a non-numeric \texttt{X} string of matching token length. Thus, the low-anchor contrast compares the low-anchor prompt to its matched control, and the high-anchor contrast compares the high-anchor prompt to its matched control.

Because multiple-choice models can be sensitive to answer-letter and position order \citep{pezeshkpour2023largelanguagemodelssensitivity}, we evaluate each item under multiple option orderings. Behavioral analyses average over 20 random option permutations per question to reduce letter- and position-order artifacts. Attribution-metric validation and circuit localization use four fixed cyclic rotations per question for computational tractability, yielding 400 paired examples per anchor contrast. We also test sensitivity to the number of answer options in Appendix~\ref{app:menu_size}.

We use four open-weight decoder-only language models from the Qwen2.5 and Llama-3.1 families \cite{Yang2024Qwen25TR,grattafiori2024llama3herdmodels}. Within each family, we compare a base model to its instruction-tuned counterpart, allowing us to examine how instruction tuning changes both behavioral anchoring and internal localization. For Qwen, we evaluate Qwen2.5-7B and Qwen2.5-7B-Instruct; for Llama, we evaluate Llama-3.1-8B and Llama-3.1-8B-Instruct. The Qwen models have 28 transformer layers, while the Llama models have 32 layers, a difference we account for when comparing localization depth. All models are run locally using the Hugging Face \texttt{transformers} library \citep{wolf2020huggingfaces}. For compactness in tables and figures, we abbreviate these models as Qwen-7B, Qwen-Inst, Llama-8B, and Llama-Inst, respectively.

\begin{table}[t]
\centering
\small
\setlength{\tabcolsep}{2.5pt}
\begin{tabular}{lccc}
\toprule
Model 
& EV shift 
& Mean \(\Delta m\) 
& \(\rho_{\mathrm{EV}}\) \\
\midrule
\multicolumn{4}{l}{\textit{Base}}\\
Qwen-7B   
& \(-0.112 / +0.134\) 
& \(-1.28 / -2.29\) 
& \(0.47 / 0.73\) \\
Llama-8B  
& \(-0.035 / +0.042\) 
& \(-0.12 / -0.31\) 
& \(0.33 / 0.75\) \\
\midrule
\multicolumn{4}{l}{\textit{Instruction-tuned}}\\
Qwen-Inst  
& \(-0.141 / +0.202\) 
& \(-3.69 / -7.10\) 
& \(0.29 / 0.60\) \\
Llama-Inst 
& \(-0.063 / +0.023\) 
& \(-0.78 / -1.41\) 
& \(0.34 / 0.18\) \\
\bottomrule
\end{tabular}
\caption{
Behavioral anchoring and attribution-metric validation. Entries are low / high. EV shift is relative to the tokenizer-matched control; Negative EV shifts indicate movement toward the low anchor, while positive shifts indicate movement toward the high anchor.  \(\Delta m\) is the anchored-minus-control change in the correct--anchor logit difference; \(\rho_{\mathrm{EV}}\) correlates \(-\Delta m\) with anchor-consistent EV shift.
}
\label{tab:behavioral_metric_validation}
\end{table}

\subsection{Behavioral and Attribution Metrics}
\label{sec:behavioral_attr_metrics}

We use behavioral metrics to measure the strength and direction of anchoring before applying circuit localization. For each item and condition, we compute the model's probability distribution over numerical candidate values by scoring the answer-label tokens, mapping label probabilities back to candidate values, and averaging over option-order permutations. We summarize anchor effects using normalized expected-value (EV) shift: the anchored-minus-control change in expected value, normalized over the candidate range, 
so negative values indicate movement toward the low anchor and positive values indicate movement toward the high anchor. We also compute total variation distance (TVD), which captures the overall size of the distributional shift. Full metric details and additional behavioral analyses are provided in Appendix~\ref{app:results_behavioral}. 
  
For circuit localization, we need a single answer-level metric whose gradients can be used to score internal nodes or edges. We use a logit difference that compares the correct answer with the answer option corresponding to the active anchor. For each item and anchor direction, let \(y_{\mathrm{correct}}\) be the option label assigned to the ground-truth value, and let \(y_{\mathrm{anchor}}\) be the option label assigned to the active low or high anchor value.

At the final answer-label prediction position, we define
\begin{equation}
\label{eq:attr_target}
m
=
\mathrm{logit}(y_{\mathrm{correct}})
-
\mathrm{logit}(y_{\mathrm{anchor}}),
\end{equation}
where \(\mathrm{logit}(y)\) denotes the model's pre-softmax score for answer-label token \(y\). Lower values of \(m\) indicate that the anchor option has become more competitive with the correct option.

For each matched prompt pair, we compute
\begin{equation}
\Delta m
=
m_{\mathrm{anchor}}
-
m_{\mathrm{control}}.
\end{equation}

\subsection{Does the Attribution Metric Track Behavioral Anchoring?}
\label{sec:metric_validation}
Before using the logit-difference metric for circuit localization, we verify two things: that the multiple-choice task elicits behavioral anchoring, and that the attribution metric tracks this behavioral effect.

\autoref{tab:behavioral_metric_validation} shows that the multiple-choice task elicits anchoring in all four models. Low anchors shift expected value downward, while high anchors shift expected value upward. The effect is strongest in the Qwen models, especially Qwen2.5-7B-Instruct, and smaller but still directionally consistent in the Llama models.

\autoref{tab:behavioral_metric_validation} also validates the attribution metric. Mean \(\Delta m\) is negative for every model and anchor contrast, indicating that anchored prompts make the anchor answer more competitive with the correct answer; all values are significant under a one-sided Wilcoxon signed-rank test (\(p<10^{-15}\)). The logit-difference shift also tracks behavioral anchoring. Because stronger anchoring corresponds to more negative \(\Delta m\), we correlate \(-\Delta m\) with anchor-consistent normalized EV shift so that larger values indicate stronger anchoring in both measures. We denote this Spearman correlation by \(\rho_{\mathrm{EV}}\). The strongest correlations appear for high-anchor contrasts in the base models, with \(\rho_{\mathrm{EV}}=0.73\) for Qwen-7B and \(0.75\) for Llama-8B, while the weakest appears for Llama-Inst on the high-anchor contrast (\(\rho_{\mathrm{EV}}=0.18\)).

We adopt \(\Delta m\) for the mechanistic analyses that follow because, although it  
does not capture every distributional effect of anchoring, 
it consistently moves in the expected direction, correlates with behavioral EV shifts, and provides an answer-level target for circuit localization.

\section{Localizing Anchor-Sensitive Pathways}

\subsection{Circuit Localization Setup}
We adopt the circuit-localization framework used in MIB~\citep{mueller2025}: a transformer is represented as a computation graph \(N\), where nodes are internal components such as attention heads or MLP modules, and edges are directed pathways between components. Under this framework, a circuit \(C \subseteq N\) is a selected subgraph of nodes or edges, and localization methods rank these nodes or edges by their estimated contribution to a target metric.

We apply this framework to anchoring. For each model and anchor direction, we use the matched prompt pairs defined in Section~\ref{sec:mc_anchoring_task}: the tokenizer-matched control prompt is the ``clean'' input, and the anchored prompt is the ``corrupted'' input. The target metric is the validated correct--anchor logit difference \(m\) from \autoref{eq:attr_target}. Since smaller \(m\) means that the anchor option has become more competitive with the correct option, we use \(L=-m\) as the attribution loss.

We compare node- and edge-level attribution patching methods following MIB. Attribution Patching (AP) approximates the effect of patching internal activations by combining the clean--corrupted activation difference with the gradient of the scalar objective. Intuitively, a node or edge receives a high score when it changes under the anchor contrast and that change affects the answer-level logit difference. We refer to node-level AP as NAP and edge-level AP as EAP. Their integrated-gradient variants, NAP-IG and EAP-IG, average attribution estimates along an interpolation path between clean and corrupted input embeddings, reducing reliance on a single local gradient. Each method returns a ranked list of candidate nodes or edges. To form a circuit, we select the top-ranked entries at a given retained fraction \(k\) and keep only entries that remain connected in the resulting subgraph before evaluating faithfulness below. Thus, \(C_k\) is formed from attribution-ranked entries rather than from a separate greedy search. For node-level methods, \(C_k\) contains the top-ranked components; for edge-level methods, \(C_k\) contains the top-ranked directed pathways.

\subsection{Recovering the Anchor-Sensitive Decision Signal}

Attribution rankings alone do not show whether the selected subgraph accounts for the control--anchored difference in \(m\). We next ask how intervening on the subgraph impacts model behavior.

The tokenizer-matched control prompt is treated as the clean input, representing the unanchored computation we want the retained circuit to recover. 
The anchored prompt is treated as the corrupted input. 
For each retained proportion \(k\), we construct a top-\(k\) circuit \(C_k\).\footnote{
\(k \in \{.001,.002,.005,.01,.02,.05,.1,.2,.5,1\}\).} 
We run the control prompt, retain the activations in \(C_k\), and replace all other activations with the corresponding activations from the matched anchored run. 
Thus, \(C_k=\emptyset\) corresponds to the fully patched anchored limit, while \(C_k=N\) corresponds to the full control run. 
Faithfulness is evaluated on \(m\), not on the negated attribution loss \(L=-m\), so recovery is measured in the same units as Section~\ref{sec:metric_validation}:
\begin{equation}
f(C_k,N;m)
=
\frac{
m(C_k)-m(\emptyset)
}{
m(N)-m(\emptyset)
}.
\label{eq:faithfulness}
\end{equation}
Here, \(m(N)\) is the full control value, and \(m(\emptyset)\) is the fully patched limit where no ranked subgraph is retained.\footnote{\(m(N)\), \(m(C_k)\), and \(m(\emptyset)\) denote mean logit-difference values over the evaluation set.} In other words, faithfulness captures the proportion of the control--anchored gap in \(m\) that is recovered by keeping \(C_k\) at its control values. Raw values can fall outside \([0,1]\) when the retained circuit undershoots or overshoots this gap.

Larger circuits generally better replicate model behavior than smaller circuits. We therefore summarize faithfulness curves with circuit performance recovery \cite[CPR;][]{mueller2025}, the area under the curve defined by \(f(C_k,N;m)\) at the various retained circuit fractions \(k\). For main-text results, faithfulness values are clipped to \([0,1]\) at each sweep point before trapezoidal integration, with endpoints added at \(k=0\) and \(k=1\) when missing. Higher CPR indicates earlier recovery with a smaller retained subgraph. Because Section~\ref{sec:metric_validation} shows that changes in \(m\) track behavioral anchoring on the same prompt pairs, we interpret faithful subgraphs as pathways relevant to the validated anchor-sensitive signal.

\subsection{Comparing Localization Methods by Faithfulness}
\label{sec:fth_results}
We first evaluate which attribution-patching method produces the most faithful circuits for the anchoring logit-difference target. This comparison determines which circuit estimate we use for the analyses that follow.

\autoref{tab:faithfulness_cpr} reports CPR for each attribution method, model, and anchor direction. 
Consistent with previous work \citep{mueller2025}, edge-level methods are 
more faithful than node-level methods: EAP consistently outperforms NAP, and EAP-IG consistently outperforms NAP-IG. This is intuitive, given that edges are more granular than nodes. Integrated gradients further improve recovery in most edge-level comparisons, with EAP-IG outperforming EAP in seven of eight model--contrast combinations. 

\begin{table}[t]
\centering
\small
\setlength{\tabcolsep}{3.5pt}
\begin{tabular}{lcccc}
\toprule
Method$\downarrow$ & Qwen-7B & Qwen-Inst & Llama-8B & Llama-Inst \\
\midrule
\multicolumn{5}{l}{\textit{Control vs. high anchor prompts}}\\
NAP    & 0.490 & 0.455 & 0.671 & 0.459 \\
EAP    & 0.923 & 0.844 & 0.946 & 0.910 \\
NAP-IG & 0.657 & 0.627 & 0.626 & 0.602 \\
EAP-IG & \textbf{0.965} & \textbf{0.966} & \textbf{0.971} & \textbf{0.953} \\
\midrule
\multicolumn{5}{l}{\textit{Control vs. low anchor prompts}}\\
NAP    & 0.361 & 0.375 & 0.636 & 0.659 \\
EAP    & 0.939 & 0.862 & \textbf{0.958} & 0.840 \\
NAP-IG & 0.705 & 0.650 & 0.714 & 0.657 \\
EAP-IG & \textbf{0.971} & \textbf{0.962} & 0.936 & \textbf{0.915} \\
\bottomrule
\end{tabular}
\caption{
Faithfulness (CPR) of attribution-patching methods. Higher values indicate a larger area-under-the-curve, and therefore recovery of the control--anchored gap in the validated logit-difference target at smaller circuit sizes. NAP/EAP denote node-/edge-level Attribution Patching; NAP-IG/EAP-IG use integrated gradients over input embeddings.
}
\label{tab:faithfulness_cpr}
\end{table}

Instruction tuning affects recoverability more strongly for EAP than for EAP-IG. For example, EAP CPR drops from 0.939 to 0.862 in Qwen on the low-anchor contrast and from 0.958 to 0.840 in Llama, while EAP-IG remains above 0.91 across all instruction-tuned settings. EAP-IG is therefore the most consistently faithful localization method in our setting.

\section{What Do EAP-IG Circuits Reveal About Anchoring Structure?}
\label{sec:eapig_circuits}

We focus our circuit analyses on EAP-IG. We first characterize where these pathways appear in the network and which component types they connect.  We then perform analyses examining the degree to which circuits are preserved across different anchor values and across base and instruction-tuned models.

\subsection{Where Are Anchoring Pathways Localized?}
\label{sec:eapig_depth_profiles}

We first ask where the estimated anchoring pathways of EAP-IG are located in the network. For each model and anchor contrast, we compute the share of total absolute EAP-IG score contributed by edges from each source layer.

\begin{figure}[t]
    \centering
    \includegraphics[height=0.28\textheight,keepaspectratio]{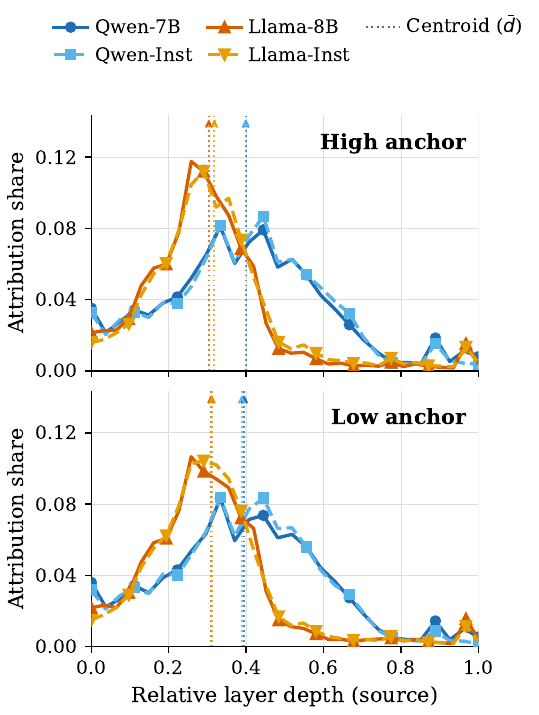}
    \caption{
    EAP-IG attribution by source-layer relative depth. For each model and anchor contrast, absolute edge scores are summed by source layer and normalized by the total absolute score. Qwen variants show a broader mid-to-late attribution profile, while Llama variants concentrate attribution earlier in the network.
    }
    \label{fig:eapig_depth_profiles}
\end{figure}

\autoref{fig:eapig_depth_profiles} shows two main layer-wise attribution patterns. First, attribution is not spread uniformly across depth. Qwen variants place more attribution in mid-to-late layers, while Llama variants concentrate attribution earlier and drop sharply after mid-depth. Second, within each model, the high- and low-anchor curves are nearly identical: the same model stays in the same depth region across anchor contrasts. The centroid lines make this visually clear, with Qwen centroids appearing later than Llama centroids in both contrasts and low/high centroids nearly overlapping within each model. Quantitatively, low- and high-anchor per-layer attribution vectors have Pearson correlations above \(0.99\), and their centroids differ by less than \(0.01\) relative depth.\footnote{Pearson correlation is computed between the low- and high-anchor vectors of per-layer attribution shares for a given model. The attribution centroid is \(\sum_{\ell} r_{\ell}a_{\ell}\), where \(r_{\ell}\) is the relative depth of layer \(\ell\) and \(a_{\ell}\) is that layer's normalized attribution share.} 
This partly aligns with prior activation-level analyses that found anchoring-related effects in earlier layers \citep{huang2026understandinganchoringeffectllm}, but shows that the localization pattern is not uniform across model families.

\begin{figure*}[t]
    \centering

    \textbf{(a) Low--high transfer within the same model}\\[-0.2em]    \includegraphics[width=0.90\linewidth,height=0.180\textheight,keepaspectratio]{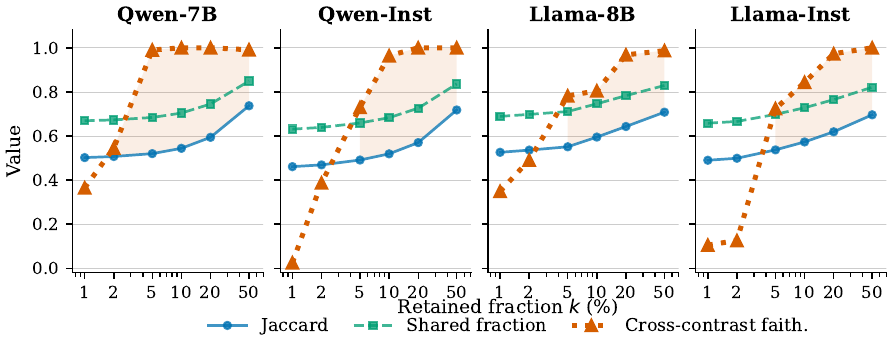}

    \vspace{0.4em}

    \textbf{(b) Base--instruction transfer within the same model family}\\[-0.2em]   \includegraphics[width=0.90\linewidth,height=0.180\textheight,keepaspectratio]{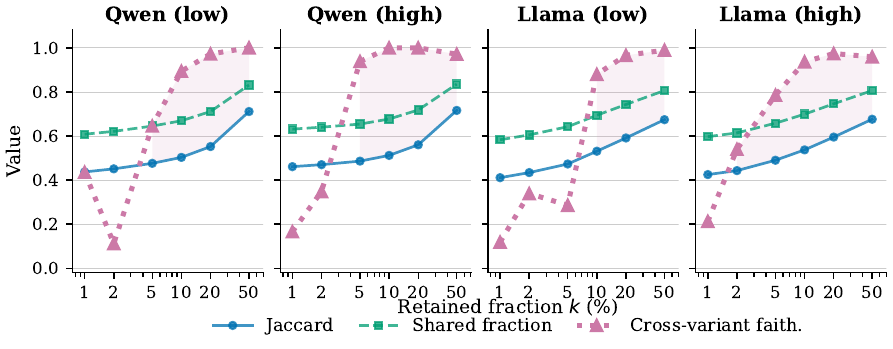}

    \vspace{-0.5em}
    \caption{
    Structural overlap and transfer of EAP-IG edge circuits. 
    Panel (a) compares low- and high-anchor circuits within each model. 
    Panel (b) compares base and instruction-tuned circuits within each model family and anchor contrast. 
    Jaccard and shared fraction measure edge overlap; faithfulness measures whether circuits replicate model behavior across the comparison. 
    }
    \label{fig:circuit_overlap_transfer}
\end{figure*}

\subsection{Which Component Types Are Involved?}
\label{sec:eapig_component_types}
Layer depth is only one way to characterize the localized pathways. We also examine which types of components are connected by the top-ranked EAP-IG edges. Most possible edges in the full graph are attention-to-attention edges, accounting for about 95--96\% of all scored edges. The top \(5\%\) circuits are also attention-heavy, but less so: attention-to-attention edges make up roughly 68--80\% of selected edges. Compared with the full graph, the selected circuits include many more attention--MLP edges. This suggests that MLPs contribute mainly through connections to and from attention components, rather than through MLP-to-MLP pathways. This pattern does not by itself identify the semantic role of these MLP-linked edges, but it suggests that MLPs participate as part of attention-mediated pathways rather than forming a separate MLP-only anchoring circuit. Full component-type breakdowns and full-graph baselines are reported in Appendix~\ref{app:other_analyses}.

\subsection{Do Low and High Anchors Use the Same Edges?}
\label{sec:low_high_overlap}
The layer-wise attribution results show that low- and high-anchor effects localize to nearly identical depth regions within each model. We next ask whether the low- and high-anchor vs. control prompt contrasts also rely on the same edges. For each model, we compare equal-sized circuits: the top \(5\%\) of EAP-IG-ranked edges for the low-anchor contrast and the top \(5\%\) for the high-anchor.

We use two simple overlap measures. Jaccard similarity, proposed in \citet{merullo_circuit_2024}, measures how much the two edge sets overlap relative to all edges selected by either contrast.\footnote{For edge sets \(A\) and \(B\), Jaccard similarity, or Intersection over Union (IoU), is \(J(A,B)=|A\cap B|/|A\cup B|\).} Because the two circuits are the same size, we also report the fraction of each circuit that is shared with the other.\footnote{For equal-sized edge sets \(A\) and \(B\), this fraction is \(|A\cap B|/|A|\), equivalently \(|A\cap B|/|B|\).} This second measure is easier to interpret: it tells us how much of a low-anchor circuit also appears in the high-anchor circuit, and vice versa.

\begin{table}[t]
\centering
\small
\setlength{\tabcolsep}{4pt}
\begin{tabular}{lcc}
\toprule
Comparison & Jaccard & Shared fraction \\
\midrule
Low--high & 0.49--0.55 & 0.66--0.71 \\
Base--instruct & 0.47--0.49 & 0.64--0.66 \\
\bottomrule
\end{tabular}
\caption{
Edge overlap between circuits at \(5\%\) retained EAP-IG edges. Ranges are across models for low--high overlap and across family and anchor-contrast pairs for base--instruction overlap.
}
\label{tab:main_overlap_summary}
\end{table}

At \(k=5\%\), \autoref{tab:main_overlap_summary} shows that low- and high-anchor circuits are not identical, but roughly two thirds of each circuit is shared with the other anchor contrast. This suggests substantial shared structure, with some edges specific to each contrast. Thus, low and high anchors differ in which individual edges rank highest, despite localizing to the same depth regions.  Full Jaccard values across retained fractions are reported in Appendix~\ref{app:other_analyses},~\autoref{tab:app_low_high_jaccard}.

\subsection{Do Base and Instruction-Tuned Models Use the Same Edges?}
\label{sec:base_instruct_overlap}

We next ask whether base and instruction-tuned variants also rely on the same top-ranked edges. For each model family and anchor contrast, we compare equal-sized circuits: the top \(5\%\) of EAP-IG-ranked edges for the base model and the top \(5\%\) for its instruction-tuned counterpart.\footnote{Qwen-7B and Qwen-Inst share the same scored edge set size, as do Llama-8B and Llama-Inst; therefore, top-\(k\%\) base--instruction circuits are equal-sized within each family.}

Table~\ref{tab:main_overlap_summary} shows that base and instruction-tuned variants share many top-ranked edges, though slightly less than the low--high circuits within a model. The shared fraction is about \(0.64\)--\(0.66\), compared with \(0.66\)--\(0.71\) for low--high overlap. This suggests that instruction tuning preserves some sparse circuit structure while changing which individual edges rank highest.

\section{Transferability of Anchoring Circuits}
\label{sec:circuit_transfer}

The structural edge overlap analyses above show that circuits can share many top-ranked edges across anchor contrasts and between base and instruction-tuned variants. However, overlap alone does not provide direct information about their functional overlap. We therefore test circuit transfer directly using faithfulness evaluations \citep{hanna2024faithfaithfulnessgoingcircuit}.

We test transfer across two axes. A \emph{matched} circuit is ranked and evaluated on the same model and anchor contrast. A \emph{cross-contrast} circuit is ranked on one control--anchor contrast and evaluated on the other within the same model: control-vs-low \(\leftrightarrow\) control-vs-high. A \emph{cross-variant} circuit is ranked on the paired base or instruction-tuned variant from the same model family, using the same anchor contrast. A \emph{cross-both} circuit changes both the anchor contrast and the model variant.

\subsection{Do Low- and High-Anchor Circuits Transfer Within a Model?}
\label{sec:low_high_transfer}
The full CPR results are reported in Appendix~\ref{app:other_analyses}, Table~\ref{tab:faithfulness_full}. Cross-contrast circuits perform almost as well as matched circuits: CPR ranges from 0.933 to 0.981 and differs from matched recovery by at most about \(0.04\). Random edge sets perform much worse, showing that transfer is not simply a consequence of retaining many edges.

\autoref{fig:circuit_overlap_transfer} complements this result by showing low--high overlap and cross-contrast faithfulness across retained fractions. At \(5\%\) retained edges, low- and high-anchor circuits have moderate Jaccard overlap but larger shared fractions, and cross-contrast faithfulness is already high. This suggests that low and high anchors do not select exactly the same edges, but enough of each circuit is shared, and those shared edges are useful enough, for circuits to transfer across anchor contrasts.

\subsection{Does Circuit Transfer Hold Across Instruction Tuning?}
\label{sec:inst_circuit_transfer}

The results above show that low- and high-anchor circuits transfer well within the same model. We next ask whether this stability also holds across instruction tuning: can a circuit found in a base model work in its instruction-tuned counterpart, and vice versa?

The full CPR results, reported in Appendix Table~\ref{tab:faithfulness_full}, show that cross-variant circuits remain well above random, indicating that instruction tuning does not create an entirely unrelated pathway. However, transfer across instruction tuning is less uniform than transfer across low- and high-anchor contrasts within the same model, especially for low-anchor contrasts and sparse retained fractions.

Figure~\ref{fig:circuit_overlap_transfer} shows this pattern. For high-anchor contrasts, cross-variant faithfulness is often close to the low--high transfer results. For low-anchor contrasts, sparse cross-variant transfer is weaker: at \(5\%\) retained edges, Qwen-7B drops from 0.917 matched faithfulness to 0.291 with the Qwen-Inst ranking, and Llama-Inst drops from 0.408 to 0.166 with the Llama-8B ranking (Appendix Table~\ref{tab:app_faithfulness_f05}). Thus, instruction tuning preserves some broader pathway structure, but the same sparse circuit does not always transfer cleanly across base and instruction-tuned variants.
 
\section{Discussion}

This work provides new insight into how language models implement numerical anchoring. Using a controlled multiple-choice anchoring task and a validated correct--anchor logit-difference metric, we apply attribution-based circuit localization to identify pathways that carry anchor-sensitive answer competition. Our results show that anchoring is better captured by sparse edge-level circuits than by isolated components. 

Within a model, low- and high-anchor circuits share substantial structure and transfer well across anchor contrasts, suggesting a shared anchor-sensitive pathway. At the same time, instruction tuning and model family shape which edges matter most and where attribution concentrates in depth. Because anchoring can shift model predictions even when the final answer remains correct, understanding these pathways is an important step toward building models that are less sensitive to irrelevant numerical context.

These findings also clarify what mitigation would need to account for. The strong low--high transfer suggests that anchoring may not need to be treated as a separate failure mode for each anchor contrast within a model. However, weaker functional transfer across base and instruction-tuned variants suggests that interventions may need to be re-checked after post-training. More broadly, applying circuit localization to cognitive biases provides a way to move beyond measuring whether a bias occurs, toward understanding how prompt-induced biases are carried through model computation.

\section*{Limitations}
\label{sec:limitations}
Our analysis uses 100 numerical-estimation questions from OpAQ. This scale is sufficient for controlled circuit-localization experiments, but it is small relative to the diversity of real-world numerical reasoning settings. Because all questions come from a single anchoring dataset, the results should be interpreted as evidence about anchoring in this controlled benchmark rather than as a comprehensive account of anchoring-like behavior in language models.

Our analysis is also specific to the multiple-choice formulation. The fixed-option setup is useful because it provides shared candidate answers, comparable answer distributions, and well-defined answer-label logits for attribution and faithfulness testing. However, it differs from open-ended numerical estimation, where models generate free-form answers and may express anchoring through different decoding dynamics, rationales, or intermediate computations. Future work should test whether the same layer-wise attribution and cross-contrast transfer results appear in open-ended settings.

We localize circuits using the correct--anchor logit difference, which captures competition between the factual answer and the answer corresponding to the active anchor at the final answer position. Although we validate that this metric tracks expected-value shifts, it does not capture the full distributional effect of anchoring. Other scalar objectives, such as KL divergence, total variation distance, expected-value shift, or anchor-side probability mass, may highlight different parts of the computation. Thus, our circuits should be understood as circuits for this validated answer-level measure of anchoring, not as a complete circuit for every way the bias can appear.

Finally, our mechanistic analysis covers four models from two families: Qwen2.5-7B, Qwen2.5-7B-Instruct, Llama-3.1-8B, and Llama-3.1-8B-Instruct. The contrast between Qwen and Llama suggests family-specific differences in where attribution concentrates across layers, but broader evaluation across additional architectures, model scales, and post-training procedures is needed before drawing general conclusions about anchoring-related circuits in language models.

\bibliography{custom}

\section*{Appendix}
\appendix

\begin{table*}[t]
\centering
\small
\setlength{\tabcolsep}{5pt}
\begin{tabular}{lcccccccc}
\toprule
Model & \(n=2\) & \(n=3\) & \(n=4\) & \(n=5\) & \(n=6\) & \(n=7\) & \(n=8\) & \(n=9\) \\
\midrule
Llama-8B & 87 & 86 & 76 & 76 & 81 & 83 & 86 & 85 \\
Llama-Inst & 38 & 40 & 36 & 42 & 45 & 52 & 57 & 58 \\
Qwen-7B & 27 & 55 & 72 & 80 & 90 & 90 & 93 & 93 \\
Qwen-Inst & 2 & 12 & 24 & 40 & 50 & 58 & 60 & 66 \\
\bottomrule
\end{tabular}
\caption{
Directional consistency across MCQA menu sizes. Values report the percentage
of items for which low anchors shift expected value downward and high anchors
shift expected value upward relative to controls.} 
\label{tab:menu_size_directional}
\end{table*}

\begin{table*}[t]
\centering
\small
\setlength{\tabcolsep}{4pt}
\begin{tabular}{lrrrr}
\toprule
Model & \(n=2\) low & \(n=2\) high & \(n=3\) low & \(n=3\) high \\
\midrule
Qwen2.5-7B-Instruct & \(+0.105\) & \(-0.421\) & \(-0.002\) & \(-0.120\) \\
Qwen2.5-7B & \(-0.091\) & \(-0.072\) & \(-0.084\) & \(+0.031\) \\
\bottomrule
\end{tabular}
\caption{
Normalized EV shifts at small menu sizes for selected models. The expected
pattern is negative for low anchors and positive for high anchors; small menus
can reverse this pattern under controls.
}
\label{tab:small_menu_signs}
\end{table*}

\section{MCQA Menu Size Validation}
\label{app:menu_size}
Our main experiments use nine answer options per item. To justify this choice,
we evaluated whether anchoring is preserved as the number of answer options
varies. For each menu size \(n \in \{2,\ldots,9\}\), we constructed MCQA
candidate sets spanning the low and high anchors and measured whether each item
showed the expected directional pattern under controls: low anchors should shift
the normalized expected value downward, while high anchors should shift it
upward.

Table~\ref{tab:menu_size_directional} reports the percentage of items showing
both expected directions. Small menus are unstable: with only two or three
options, several models fail to recover the expected bidirectional anchoring
pattern. In some cases, especially for Qwen instruction-tuned models, the
normalized expected-value shifts even have the wrong sign. Directional
consistency improves as the menu size increases and stabilizes around
\(n \geq 5\)--\(7\). The nine-option setting used in the main experiments is
therefore not arbitrary: it is the largest tested menu size and is either the
maximum or within a few percentage points of the maximum directional consistency
for all models.

\section{Behavioral Validation}
\label{app:results_behavioral}
Before localizing circuits, we first verify that our MCQA transformation
preserves the behavioral signature of anchoring. This step is necessary
because the mechanistic analysis relies on a multiple-choice format, whereas
anchoring is typically studied in open-ended numerical estimation. 

\begin{table*}[t]
\centering
\small
\setlength{\tabcolsep}{5pt}
\begin{tabular}{l cc cc cc}
\toprule
& \multicolumn{2}{c}{EV shift} 
& \multicolumn{2}{c}{Directional adherence} 
& \multicolumn{2}{c}{TVD} \\
\cmidrule(lr){2-3} \cmidrule(lr){4-5} \cmidrule(lr){6-7}
Model 
& $\Delta_{\mathrm{low}} \downarrow$ 
& $\Delta_{\mathrm{high}} \uparrow$
& Low 
& High 
& Low 
& High \\
\midrule
Llama-3.1-8B 
& $-0.035$ & $+0.042$ & 98 & 87 & 0.061 & 0.097 \\
Llama-3.1-8B-Instruct 
& $-0.063$ & $+0.023$ & 92 & 65 & 0.154 & 0.160 \\
Qwen2.5-7B 
& $-0.112$ & $+0.134$ & 98 & 95 & 0.198 & 0.261 \\
Qwen2.5-7B-Instruct 
& $-0.141$ & $+0.202$ & 75 & 82 & 0.369 & 0.442 \\
\bottomrule
\end{tabular}
\caption{
Behavioral anchoring effects after MCQA conversion. 
EV shifts are relative to controls, adherence counts expected-sign shifts, and TVD measures distributional change.
}
\label{tab:behavioral_anchoring}
\end{table*}

\paragraph{Behavioral Metrics}
We use behavioral metrics from \citet{owusu2026anchoring} to verify that the multiple-choice task preserves the anchoring effect studied in prior behavioral work. These metrics validate the task setup, while the circuit analyses focus on the correct--anchor logit-difference metric. In this
setting, anchoring means that an irrelevant numerical prime shifts the
model's probability distribution over candidate answers: low anchors should
move probability mass toward smaller values, while high anchors should move
probability mass toward larger values. We therefore measure both the direction
of the shift, using expected value, and the overall size of the distributional
change, using total variation distance.

For each item \(i\), let \(\mathcal{L}=\{A,\ldots,I\}\) denote the set of
answer labels, and let \(\mathcal{V}_i\) denote the corresponding set of
numerical candidate values. For a prompt condition \(c\), the model assigns a
logit \(z_i^{(c)}(\ell)\) to each answer label \(\ell \in \mathcal{L}\) at the
final answer-label prediction position. We obtain a probability distribution
over answer labels by applying a softmax over the nine label logits:
\begin{equation}
p_i^{(c)}(\ell)
=
\frac{\exp z_i^{(c)}(\ell)}
{\sum_{\ell' \in \mathcal{L}} \exp z_i^{(c)}(\ell')}.
\end{equation}

Because answer labels are randomly permuted across trials, we compute label
probabilities separately for each permutation, map those probabilities back to
their associated numerical values, and then average the resulting
value-space distributions across permutations. This yields
\(q_i^{(c)}(v)\), the permutation-averaged probability assigned to numerical
candidate \(v \in \mathcal{V}_i\). Thus, \(q_i^{(c)}\) is a distribution over
candidate values, not over fixed answer labels or display positions.

We summarize the model's distribution under condition \(c\) by its expected
value:
\begin{equation}
\mathrm{EV}_i^{(c)}
=
\sum_{v \in \mathcal{V}_i} q_i^{(c)}(v) v .
\end{equation}

For each anchor direction \(d \in \{\mathrm{low}, \mathrm{high}\}\), let
\(b(d)\) denote the matched control baseline for that direction. The
low-anchor prompt is compared to a control prompt matched to the low-anchor
string, and the high-anchor prompt is compared to a control prompt matched to
the high-anchor string. Both controls preserve the slot-machine sentence but
replace the numerical anchor with a non-numeric \texttt{X} string matched to
the tokenizer length of the corresponding anchor. In some examples these two
matched controls are identical, but we define them separately because low and
high anchor strings need not tokenize the same way.

We define the normalized expected-value shift as
\begin{equation}
\Delta \mathrm{EV}_i^{(d)}
=
\frac{
\mathrm{EV}_i^{(d)} - \mathrm{EV}_i^{(b(d))}
}{
v_i^{\max} - v_i^{\min}
}.
\end{equation}
Negative values indicate shifts toward lower estimates, while positive values
indicate shifts toward higher estimates. Thus, anchoring predicts
\(\Delta \mathrm{EV}_i^{(\mathrm{low})}<0\) and
\(\Delta \mathrm{EV}_i^{(\mathrm{high})}>0\).

We also measure the total variation distance between the anchored and matched
control distributions:
\begin{equation}
\mathrm{TVD}_i^{(d)}
=
\frac{1}{2}
\sum_{v \in \mathcal{V}_i}
\left|
q_i^{(d)}(v) - q_i^{(b(d))}(v)
\right|.
\end{equation}
Unlike \(\Delta \mathrm{EV}\), TVD is non-directional: it measures the
magnitude of the distributional change regardless of whether the expected
value shifts upward or downward.

\paragraph{Results}

Normalized expected-value shifts test whether the model's answer distribution
moves in the expected direction, while TVD measures the magnitude of the
distributional change regardless of direction. A preserved anchoring effect
predicts $\Delta_i^{(\mathrm{low})}<0$ for low anchors and
$\Delta_i^{(\mathrm{high})}>0$ for high anchors.

Table~\ref{tab:behavioral_anchoring} shows that the MCQA transformation
preserves the expected directionality 
of anchoring across all four
models. Mean normalized shifts are negative for low anchors and positive for
high anchors in every model, indicating that low anchors pull the expected
answer downward while high anchors pull it upward relative to controls.

Llama-3.1-8B moves in the expected direction on 98/100 low-anchor questions and 87/100
high-anchor questions, with mean shifts of $-0.035$ and $+0.042$,
respectively. Qwen2.5-7B shows larger aggregate shifts
($-0.112$ for low anchors and $+0.134$ for high anchors) and high directional
consistency in both directions, moving as expected on 98/100 low-anchor and
95/100 high-anchor questions.

\section{EAP-IG Circuit Analyses}
\label{app:other_analyses}

\paragraph{Component composition.}
Table~\ref{tab:app_component_composition} compares the component-type composition of the full scored edge set with the top \(5\%\) EAP-IG circuits. Most possible edges in the full graph are attention-to-attention edges, but the selected circuits contain proportionally more attention--MLP edges. This suggests that MLPs contribute mainly through connections to and from attention components rather than through MLP-to-MLP pathways.

\paragraph{Low--high overlap across retained fractions.}
Table~\ref{tab:low_high_overlap_crossfaith} compares structural overlap and functional transfer between low- and high-anchor EAP-IG circuits. Jaccard overlap is moderate, but the shared fraction shows that a large portion of each equal-sized circuit appears in the other. Cross-contrast faithfulness rises quickly as more top-ranked edges are retained, indicating that the two anchor contrasts are not identical as edge sets but are functionally related.

\paragraph{Sparse-circuit transfer.}
Table~\ref{tab:app_faithfulness_f05} reports pointwise faithfulness at \(5\%\) retained EAP-IG edges. Unlike CPR, which summarizes recovery across the full retained-fraction curve, this table tests whether transfer holds at a sparse circuit size. The results show that low--high transfer is often strong even at \(5\%\), while cross-variant transfer is more variable, suggesting that instruction tuning changes which sparse edges matter most.

\paragraph{Additional transfer results.}
Tables~\ref{tab:app_faithfulness_f05} and~\ref{tab:faithfulness_full} provide complementary transfer results for EAP-IG edge circuits. Table~\ref{tab:app_faithfulness_f05} reports pointwise faithfulness at \(5\%\) retained edges, while Table~\ref{tab:faithfulness_full} reports CPR across the full retained-fraction sweep.

\begin{table*}[t]
\centering
\small
\setlength{\tabcolsep}{4pt}
\begin{tabular}{llrrrrrrr}
\toprule
Set & Model & A$\to$A & A$\to$M & M$\to$A & M$\to$M & E$\to$A & E$\to$M & $\to$logit \\
\midrule
Full graph & Qwen-7B    & 95.0 & 1.2 & 3.4 & 0.0 & 0.3 & 0.0 & 0.1 \\
High top 5\% & Qwen-7B & 68.7 & 10.6 & 19.6 & 0.6 & 0.1 & 0.0 & 0.3 \\
Low top 5\%  & Qwen-7B & 68.0 & 11.0 & 19.8 & 0.7 & 0.1 & 0.0 & 0.4 \\
\addlinespace[2pt]

Full graph & Qwen-Inst  & 95.0 & 1.2 & 3.4 & 0.0 & 0.3 & 0.0 & 0.1 \\
High top 5\% & Qwen-Inst & 69.5 & 10.4 & 19.0 & 0.6 & 0.1 & 0.0 & 0.3 \\
Low top 5\%  & Qwen-Inst & 69.2 & 10.7 & 19.1 & 0.6 & 0.1 & 0.0 & 0.2 \\
\addlinespace[2pt]

Full graph & Llama-8B   & 95.7 & 1.1 & 3.0 & 0.0 & 0.2 & 0.0 & 0.1 \\
High top 5\% & Llama-8B & 80.3 & 6.5 & 11.7 & 0.5 & 0.5 & 0.0 & 0.5 \\
Low top 5\%  & Llama-8B & 79.5 & 6.9 & 12.1 & 0.5 & 0.4 & 0.0 & 0.6 \\
\addlinespace[2pt]

Full graph & Llama-Inst & 95.7 & 1.1 & 3.0 & 0.0 & 0.2 & 0.0 & 0.1 \\
High top 5\% & Llama-Inst & 78.8 & 7.3 & 12.5 & 0.5 & 0.4 & 0.0 & 0.5 \\
Low top 5\%  & Llama-Inst & 79.5 & 7.0 & 12.2 & 0.5 & 0.4 & 0.0 & 0.4 \\
\bottomrule
\end{tabular}
\caption{
Component-type composition of the full scored edge set and the top \(5\%\) EAP-IG edge circuits by absolute attribution score. Values are percentages of edges. A = attention component, M = MLP component, and E = embedding. The full graph rows show the component-type distribution before ranking by EAP-IG score.
}
\label{tab:app_component_composition}
\end{table*}

\label{app:jacc_full}
\begin{table*}[t]
\centering
\small
\setlength{\tabcolsep}{4pt}
\begin{tabular}{lcccccc}
\toprule
Model & 1\% & 2\% & 5\% & 10\% & 20\% & 50\% \\
\midrule
Qwen-7B    & 0.502 & 0.507 & 0.520 & 0.544 & 0.594 & 0.737 \\
Qwen-Inst  & 0.461 & 0.469 & 0.492 & 0.519 & 0.570 & 0.719 \\
Llama-8B   & 0.526 & 0.536 & 0.551 & 0.595 & 0.643 & 0.708 \\
Llama-Inst & 0.490 & 0.499 & 0.537 & 0.573 & 0.619 & 0.696 \\
\bottomrule
\end{tabular}
\caption{
Jaccard similarity between low- and high-anchor EAP-IG edge circuits across retained fractions. For each model and retained fraction \(k\), we compare the top-\(k\)\% edges ranked by absolute EAP-IG score for the low-anchor contrast and the high-anchor contrast. Higher values indicate greater overlap between the two edge sets.
}
\label{tab:app_low_high_jaccard}
\end{table*}

\begin{table*}[t]
\centering
\small
\setlength{\tabcolsep}{4pt}
\begin{tabular}{llccccc}
\toprule
Eval. contrast & Model & Matched & Cross-contrast & Cross-variant & Cross-both & Random \\
\midrule
\multirow{4}{*}{Low}
& Qwen-7B    & 0.971 & 0.969 & 0.910 & 0.932 & 0.257 \\
& Qwen-Inst  & 0.962 & 0.972 & 0.976 & 0.969 & 0.250 \\
& Llama-8B   & 0.936 & 0.933 & 0.909 & 0.935 & 0.328 \\
& Llama-Inst & 0.915 & 0.950 & 0.939 & 0.890 & 0.250 \\
\midrule
\multirow{4}{*}{High}
& Qwen-7B    & 0.965 & 0.981 & 0.964 & 0.937 & 0.265 \\
& Qwen-Inst  & 0.966 & 0.947 & 0.955 & 0.981 & 0.253 \\
& Llama-8B   & 0.971 & 0.954 & 0.985 & 0.916 & 0.251 \\
& Llama-Inst & 0.953 & 0.931 & 0.903 & 0.896 & 0.251 \\
\bottomrule
\end{tabular}
\caption{
Cross-condition faithfulness of EAP-IG edge circuits.
Matched uses circuits ranked on the same model and anchor contrast as evaluation. 
Cross-contrast uses the same model but the opposite anchor contrast. 
Cross-variant uses the paired base or instruction-tuned variant from the same model family with the same anchor contrast. 
Cross-both changes both the model variant and anchor contrast. 
Random uses same-size random edge sets.
}
\label{tab:faithfulness_full}
\end{table*}

\begin{table*}[t]
\centering
\small
\setlength{\tabcolsep}{4pt}
\begin{tabular}{llcccccc}
\toprule
Model & Measure & 1\% & 2\% & 5\% & 10\% & 20\% & 50\% \\
\midrule
\multirow{3}{*}{Qwen-7B} 
& Jaccard & 0.502 & 0.507 & 0.520 & 0.544 & 0.594 & 0.737 \\
& Shared fraction & 0.669 & 0.673 & 0.684 & 0.704 & 0.745 & 0.849 \\
& Cross-contrast faith. & 0.365 & 0.545 & 0.990 & 1.000 & 1.000 & 0.991 \\
\addlinespace[2pt]
\multirow{3}{*}{Qwen-Inst} 
& Jaccard & 0.461 & 0.469 & 0.491 & 0.519 & 0.570 & 0.718 \\
& Shared fraction & 0.631 & 0.639 & 0.659 & 0.683 & 0.726 & 0.836 \\
& Cross-contrast faith. & 0.025 & 0.387 & 0.730 & 0.964 & 1.000 & 1.000 \\
\addlinespace[2pt]
\multirow{3}{*}{Llama-8B} 
& Jaccard & 0.526 & 0.536 & 0.551 & 0.595 & 0.643 & 0.708 \\
& Shared fraction & 0.689 & 0.698 & 0.711 & 0.746 & 0.783 & 0.829 \\
& Cross-contrast faith. & 0.349 & 0.491 & 0.782 & 0.805 & 0.968 & 0.987 \\
\addlinespace[2pt]
\multirow{3}{*}{Llama-Inst} 
& Jaccard & 0.490 & 0.499 & 0.537 & 0.573 & 0.619 & 0.696 \\
& Shared fraction & 0.658 & 0.666 & 0.698 & 0.728 & 0.765 & 0.821 \\
& Cross-contrast faith. & 0.105 & 0.126 & 0.724 & 0.843 & 0.973 & 1.000 \\
\bottomrule
\end{tabular}
\caption{
Low--high circuit overlap and cross-contrast faithfulness across retained fractions for EAP-IG edge circuits. Jaccard measures overlap relative to the union of the low- and high-anchor edge sets. Shared fraction measures how much of each equal-sized top-\(k\) circuit is shared with the other. Cross-contrast faithfulness reports average faithfulness when a circuit ranked on one control--anchor contrast is evaluated on the other, with each direction clipped to \([0,1]\). These measures separate edge-set overlap from functional transfer.
}
\label{tab:low_high_overlap_crossfaith}
\end{table*}
\

\begin{table*}[t]
\centering
\small
\setlength{\tabcolsep}{4pt}
\begin{tabular}{llccccc}
\toprule
Eval. & Model & Matched & Cross-contrast & Cross-variant & Cross-both & Random \\
\midrule
\multicolumn{7}{l}{\textit{Low-anchor evaluation}}\\
Low & Qwen-7B    & 0.917 & 0.989 & 0.291 & 0.777 & 0.000 \\
Low & Qwen-Inst  & 0.890 & 1.743 & 1.863 & 1.044 & 0.000 \\
Low & Llama-8B   & 0.678 & 0.689 & 0.405 & 0.667 & -0.009 \\
Low & Llama-Inst & 0.408 & 0.807 & 0.166 & 0.540 & -0.001 \\
\midrule
\multicolumn{7}{l}{\textit{High-anchor evaluation}}\\
High & Qwen-7B    & 0.998 & 0.991 & 1.103 & 0.442 & 0.000 \\
High & Qwen-Inst  & 1.066 & 0.461 & 0.875 & 1.038 & 0.000 \\
High & Llama-8B   & 0.856 & 0.874 & 1.000 & 0.561 & 0.000 \\
High & Llama-Inst & 0.864 & 0.641 & 0.571 & 0.568 & 0.000 \\
\bottomrule
\end{tabular}
\caption{
Sparse-circuit transfer at \(5\%\) retained EAP-IG edges. Values are pointwise unclipped faithfulness values at the retained-fraction sweep point nearest \(5\%\) (unlike CPR). Matched uses circuits ranked on the same model and anchor contrast as evaluation. Cross-contrast uses the same model but the opposite anchor contrast. Cross-variant uses the paired base or instruction-tuned variant from the same model family with the same anchor contrast. Cross-both changes both the model variant and anchor contrast. Random uses same-size random edge sets. Values are unclipped and may exceed 1 when the retained circuit overshoots the full-model recovery.
}
\label{tab:app_faithfulness_f05}
\end{table*}

\begin{table*}[t]
\centering
\small
\setlength{\tabcolsep}{4pt}
\begin{tabular}{lccc}
\toprule
Model & Jaccard & Shared frac. & Cross-contrast faith. \\
\midrule
Qwen-7B    & 0.520 & 0.684 & 0.990 \\
Qwen-Inst  & 0.491 & 0.659 & 0.730 \\
Llama-8B   & 0.551 & 0.711 & 0.782 \\
Llama-Inst & 0.537 & 0.698 & 0.724 \\
\bottomrule
\end{tabular}
\caption{
Low--high edge overlap and transfer at \(5\%\) retained EAP-IG edges. Jaccard measures overlap relative to the union of the two edge sets; shared fraction measures how much of each equal-sized circuit is shared; cross-contrast faithfulness averages low-to-high and high-to-low faithfulness, clipped to \([0,1]\).
}
\label{tab:low_high_overlap_5pct}
\end{table*}

\end{document}